\documentclass[11pt,letterpaper]{article}

\usepackage[margin=1in]{geometry}
\usepackage[T1]{fontenc}
\usepackage[utf8]{inputenc}
\usepackage{lmodern}
\usepackage{microtype}
\usepackage{booktabs}
\usepackage{graphicx}
\usepackage{tablefootnote}
\usepackage{longtable}
\usepackage{caption}
\usepackage{amssymb} 
\usepackage{titling}
\usepackage[backend=biber,style=authoryear,natbib=true]{biblatex}
\graphicspath{{./images/}}

\title{NightmareDiscourseBench: A Benchmark Dataset for Fictional Direct Discourse Identification Pipelines\\[0.5em]
\large\textnormal{Preprint}}
\author{Claudia Carroll\thanks{Department of English, Carnegie Mellon University, Pittsburgh, PA, USA. ORCID: 0000-0002-1853-9106}
\and Gabi Kirilloff\thanks{Department of English, Carnegie Mellon University, Pittsburgh, PA, USA. ORCID: 0000-0003-2696-1419}
\and Dashiell Leigh\thanks{Washington University in St. Louis, St. Louis, MO, USA}}
\date{} 

\begin{document}

\maketitle

\begin{abstract}
In this paper we introduce a new benchmark dataset, NightmareDiscourseBench, for testing NLP pipelines that identify forms of direct discourse in English-language fiction. We first review existing publicly available benchmarks (providing summary statistics and reviewing available metadata) for this task, identifying theoretical inconsistency and exclusive focus on speech as primary issues in this domain.
We then overview our narrative theory-based framework for classifying and evaluating different forms of direct discourse, and finally describe NightmareDiscourseBench, a new benchmark test set of 12 text samples accompanied by metadata pertinent to the task of direct discourse identification.
\end{abstract}

\section{Introduction}  

In this paper we introduce a new benchmark dataset, NightmareDiscourseBench, for testing NLP pipelines that identify forms of direct discourse in English-language fiction. Direct discourse, an umbrella category for the parts of fictional text that are the verbatim report of words of entities other than the primary narrator (usually in the form of direct speech, or dialogue), is of interest to many literary critics because of its contribution to central elements of fictional form–including the representation of social interaction, characters’ differing perspectives and communication strategies, the analysis of fictional social networks, and issues of whose voices get foregrounded. Additionally, the accurate identification and removal of direct discourse is also necessary for research on narrator representation, such as the use of direct address and narrator commentary. We first overview our narrative theory-based framework for classifying and evaluating different forms of direct discourse, review existing publicly available benchmarks (providing summary statistics and reviewing available metadata), and finally describe NightmareDiscourseBench, a new benchmark test set of 12 text samples accompanied by metadata pertinent to the task of direct discourse identification–such as narrative perspective, typography, and place of publication. 

NightmareDiscourseBench is designed as a stress test dataset and contains text samples that are particularly challenging from the perspective of direct discourse identification. It provides advantages over existing test sets in this domain not only by focusing on challenging text samples, but also by providing data annotations for three kinds of direct discourse–speech, thought and embedded text–as well as an ‘other’ category for ambiguous cases that may challenge identification pipelines. We finally demonstrate the utility of NightmareDiscourseBench by comparing the accuracy of a widely used NLP tool for direct discourse identification, BookNLP, on the three existing benchmarks and NightmareDiscourseBench.

Identification of direct discourse is a common task in computational literary studies. Direct discourse, a category which includes the words of fictional characters in the form of direct speech, direct thought, or embedded materials such as letters, songs and quotations, is usually either an object of study in itself (for researchers focused on the voices of fictional characters), or necessitates removal or separation from indirect discourse (or narration) for researchers focused on narratorial description, representation or commentary. For example, researchers have studied fictional speech patterns in English \cite{menon2019} and German \cite{Jannidis2018AnalysingDS} novels, analyzed plot arcs as measured by sentiment in direct speech \cite{vishnubhotla2024emotion}, and used direct speech as a proxy for social networks. \cite{sims2020measuring, kubis2021quantitative, lee2024conversational} A closely related area of research that requires direct discourse identification is speaker identification–the connection of each instance of direct speech in a novel to the character speaking it. \cite{elson2010extracting, he2013identification, vishnubhotla2022dialogism, michel2024distinguishing, michel2024improving, lee2016annotated}

While the identification or removal of direct discourse (or one of its subcategories) is a common task in computational literary studies, it is often treated as a trivial one in computational research on English-language fiction. For researchers focused on English language texts, identification of direct discourse, particularly direct speech, is often treated as a simple matter of identifying text enclosed within quotation marks. \cite{elson2010extracting, menon2019, sims2020measuring, vishnubhotla2024emotion, kubis2021quantitative} In such cases, direct discourse identification is framed as basic data cleaning, and researchers often provide minimal to no information on their identification pipelines, test sets, or accuracy rates. Because of this assumption of triviality, English-language benchmark datasets for the training or testing of automated direct discourse identification pipelines are few, and often under-documented by their creators or not shared publicly. However, even for many English-language fiction corpora the process of direct discourse identification (even in its most simple form—direct speech) is in fact quite challenging. The accuracy rates of quotation mark-based identification pipelines can vary dramatically depending on date of publication, region of publication, typographical standards and even the edition of a particular text. This assumption of triviality is in contrast to research on fiction in other languages, for which the absence of a standard typographical means of delimiting quotation marks have forced researchers to be more creative. The main approach to direct discourse identification that does not rely on rules-based extraction via quotation marks is binary token classification using machine learning models. However, these approaches have primarily been tested on non-English texts, including Danish \cite{allaith2025annotating}, Norwegian \cite{allaith2025annotating}, Swedish \cite{stymne2022slanda, stymne2024direct}, German \cite{brunner2013automatic} and French \cite{schoch2016straight} with two examples of multi-lingual direct discourse identification using BERT. \cite{byszuk2020detecting, kurfali2020zeroshot}

The conceptual category of direct discourse is also often under-theorized in computational research on the topic. There are several forms of direct discourse, and common subcategories like direct speech, direct thought and embedded writing may or may not be encased in quotation marks, depending on the context. Related categories like quotation, dialogue and speech are often used interchangeably or equated with each other in both identification pipelines and training/test datasets. Therefore, for researchers aiming to identify direct discourse or its subcategories in their fiction corpora, it is often 1) difficult to judge the applicability of available identification pipelines and training/test sets for their particular research needs and 2) necessary to build and test a pipeline from scratch in order to ensure compatibility between the research question and processed text data.

\section{Theoretical Framework} 

In narrative theory, direct discourse is one of the three primary speech-discourse categories for the representation of the experience of the characters: indirect discourse, free indirect discourse and direct discourse. The speech discourse approach frames narrative as a primary speech act (the utterance of the narrator) and a set of embedded speech acts by diegetic characters or other entities (where “speech” is used broadly to refer to words spoken, thought, written or otherwise communicated by an entity). The various forms of diegetic speech-discourse are all framed by the narration, the words “spoken” by the narrator in the telling of the story. Direct discourse occurs when “a character’s utterances or thoughts are given in or quoted in the way the character (presumably) formulated them”. \cite{prince} As Brian McHale explains, “[Direct discourse] is conventionally understood to replicate exactly what the quoted character is supposed to have said or thought, preserving (for instance) expressive elements of the original utterance.” McHale adds that “in direct discourse (DD), whether it represents a speech event or an unspoken thought, the transition from frame to inset is clearly visible, typically signaled typographically and/or by an introductory verb of speech or thought: “She said,” “She thought.”) \cite{mchale2009}

We recognize that, in practice, the majority of the text enclosed in quotation marks in English-language fiction is direct speech. However, we find that the terminological slippage can 1) result in the reduction of categories like direct thought and embedded text to errors or edge-cases when they are forms of character representation of interest in themselves, and 2) disguise how pipelines for the identification of direct speech can perform dramatically worse on certain texts, because of their above-average use of such representational techniques. While, as McHale notes, direct discourse is usually delimited by typographical signals, speech or thought tags, or both, it does not have to be. Samuel Richardson’s 1740 novel \textit{Pamela}, for example, reports speech exclusively using speech tags, without any accompanying typographical signals:

\begin{quote}
His waistcoat stood on end with silver lace, and he looked very grand.  But what he did last, has made me very serious, and I could make him no compliments. Said he, Why don’t you wear your usual clothes? Though I think every thing looks well upon you (for I still continue in my new dress). I said, I have no clothes, sir, I ought to call my own, but these: and it is no matter what such an one as I wears. \cite{richardson2004}
\end{quote}

In this case, a quotation mark based identification pipeline will miss every piece of dialogue or speech in the text. Furthermore, not everything delimited by typographical markers is direct discourse. Another common source of error in direct speech identification pipelines that rely on typographical markers are scare quotes, often used to indicate irony on the part of the narrator rather than direct discourse. Take, for example, a passage from the opening paragraph of Wharton’s \textit{Age of Innocence}:

\begin{quote}
There was no reason why the young man should not have come earlier, for he had dined at seven, alone with his mother and sister, and had lingered afterward over a cigar in the Gothic library with glazed black-walnut bookcases and finial-topped chairs which was the only room in the house where Mrs. Archer allowed smoking. But, in the first place, New York was a metropolis, and perfectly aware that in metropolises it was “not the thing” to arrive early at the opera; and what was or was not “the thing” played a part as important in Newland Archer's New York as the inscrutable totem terrors that had ruled the destinies of his forefathers thousands of years ago. \cite{wharton}
\end{quote}

While the text enclosed in quotation marks is minimal in this passage, and therefore unlikely to significantly affect analysis of the novel’s use of direct speech at scale, its accurate identification as the words of the narrator, and not that of the characters, is necessary for analysis of narratorial commentary and irony.  Direct discourse, or even its most common form, direct speech, should therefore not be equated with text enclosed in quotation marks. (The most common challenges associated with direct discourse identification pipelines that rely on quotation heuristics are listed, with examples, in the Appendix (Table 5).) Use of speech/thought tags to identify direct speech or thought is also unreliable–speech/thought tags are a characteristic that also mark indirect discourse, in which the words or thoughts of a character are summarized by the narrator (e.g. he thought that the shirt was ugly.) Finally, the other commonly used term to describe this form of discourse, dialogue, is also conceptually unsuitable–while most direct speech is likely dialogue in English language fiction, dialogue is an instance of direct speech that is directed toward, or in response to, another diegetic entity. Direct speech that is self-directed does not fall under this category. Currently, quotation, dialogue and direct speech (or simply “speech”) are all used largely interchangeably in English-language computational literary studies work on direct discourse. A more granular theoretical framing of direct discourse identification can assist scholars in determining whether certain pipelines are suitable for their particular text corpora and research question. 

The key characteristic of direct discourse is the verbatim report of the words of the character, be it their speech, thoughts or other forms of communication. On this basis, we frame our literature review and our own benchmark dataset using three primary categories of direct discourse:

\begin{enumerate}
    \item Direct speech: The most common form of direct discourse, in which the spoken words of the character are rendered verbatim by the narrator. Direct speech may occur when the character is in conversation with other characters (i.e. dialogue) or when the character speaks to themselves.
    \item Direct thought: Cases when the thoughts of a character are reported verbatim by the narrator. 
    \item Embedded text: Cases where the written words of a character are reported verbatim by the narrator. Common examples include letters, and direct quotation of textual works from either the real world or the storyworld, such as newsletters, advertisements and famous literary quotations.
\end{enumerate}

Embedded written text produced by characters is not always included in definitions of direct discourse. We include it here because 1) such text is distinct from the narration in that it is the words of the characters reported verbatim and 2) such text is often enclosed in quotation marks, or shared other lexical/stylistic features with direct speech and thought, such that it is a common false positive in automated pipelines aiming to identify direct speech.

\section{Existing Benchmark Review} 

With the above categorical framing in mind, in the next section we review the three publicly available English-language datasets to train or test automated identification of direct discourse or one of its subcategories–LitBank, the Quote Annotator test set, and RiQuA. Note that these are not all the datasets we identified in our review, just those that are publicly available. We do not include here hand-annotated datasets that are mentioned in published research but were never published by authors \cite{menon2019, byszuk2020detecting} or datasets that were published, but for which the links are dead. \cite{brooke2017lexical} We also do not include datasets for which the fictionality is ambiguous, such as religious texts \cite{lee2016annotated,lee2024conversational}, or narrative history. \cite{hess2020narrator} Finally, we also do not include datasets which are often used as sources of direct discourse data, but which were not produced by hand annotation, such as the Columbia Quoted Speech Attribution Corpus \cite{elson2010extracting} and the Project Dialogism Novel Corpus. \cite{vishnubhotla2022dialogism}

\subsection{LitBank}

The LitBank annotated dataset is the largest available dataset of literary text hand-annoted for a form of direct discourse. \cite{bamman2019litbank}\footnote{https://github.com/dbamman/litbank} The LitBank annotations include hand-annotated instances of dialogue from the first ~2000 words of 86 novels, with 75 unique authors. (LitBank contains annotations for other categories, and contains a total of 100 text samples, however 14 of these do not contain any dialogue in their first 2000 words). All texts are in English, are primarily British or American (with a handful of Anglo-Irish and Australian texts included), have publication dates that range between 1719 and 1922, and are sourced from Project Gutenberg. Annotations were created by “four annotators after a period of initial training, prompted to identify all quotations and attribute each one to the speaker who uttered it.” \cite{sims2020measuring} The strengths of the LitBank dataset for the purpose of direct speech identification evaluation is its range–it contains the most annotated novel segments, by far, and therefore has the greatest variety of authors and novels out of all available benchmarks in this domain.

However, the LitBank annotations have two key weaknesses in respect to direct discourse identification. First, the samples taken from each of the 100 novels consists of the first ~2000 tokens, rounded to the nearest whole sentence. While sampling the first part of each novel reduces the biases involved in manual selection, and makes the samples more directly comparable amongst themselves, the opening of a novel is often more formally tidy than the rest of the novel as the author introduces new characters, settings and plot points to the reader. Furthermore, there is evidence to suggest that the opening of a novel has, on average, the lowest concentration of direct discourse compared to the middle and end (and indeed, the relative frequency of text spans tagged as dialogue in the LitBank dataset is less than half that of the other benchmark dataset in this domain). \cite{menon2019} Second, the description of the LitBank dataset, and associated papers, varies in how it describes the object of annotation within the realm of direct discourse. The dataset hosted on GitHub, for example, describes the annotations as both “quotations” and “direct speech”. The paper linked in the same section suggests that the target object for annotation was “dialogue”, and notes that false positives in their test run were largely text enclosed in quotations “including scare quotes for emphatic use (to introduce jargon, neologisms, or irony), titles of works of art, the mention of a term (as distinct from its use), and written use.” \cite{sims2020measuring}  Within the gold label dataset itself, there are additional small conceptual inconsistencies–for example, the words of a speaking parrot in The Awakening, and Selected Short Stories is not included in the gold annotation set, despite arguably fitting the accepted criteria for both direct speech and dialogue (the parrot is speaking to, or at least at, a main character). The annotation guidelines used by the four annotators are not provided. The LitBank quotation annotations are therefore best described as annotations of direct speech enclosed in quotation marks not including self-directed character speech.

\subsection{Quote Annotator}

Muzny et al.’s \cite{Muzny2017a} Quote Annotator evaluation set contains six full length novels hand annotated for quotation boundaries.\footnote{The full dataset is available at https://muzny.github.io/dialogism.html} The novels are all in English, published between 1788 and 2011, and include three British (Charlotte Smith’s \textit{Emmeline the Orphan}, Austen’s \textit{Pride and Prejudice}, and E.M. Forster’s \textit{A Room With A View} and three US novels. (James Fenimore Cooper’s \textit{The Spy}, F. Scott Fitzgerald’s \textit{The Great Gatsby}, and John Grisham’s \textit{The Litigators}) In contrast to the LitBank dataset, the strength of this dataset as a benchmark for direct discourse identification is its thoroughness–while diversity is lost by focusing on only six novels, the annotations cover each novel in full, thereby offering a range of direct discourse forms across narrative time. The primary weakness of this dataset as a benchmark is, like LitBank, the fuzziness of its conceptual boundaries. Muzny et al. describe the NLP pipeline for which the dataset was designed as one aiming to extract “dialogue—instances of direct quoted speech”. \cite{Muzny2017a} Dialogue and direct speech (while, again largely overlapping) are not synonymous. Additionally, the dataset actually includes annotations of all text spans in these six novels enclosed in quotation marks. Quotations with no identifiable speaker, including quotations from embedded text, scare quotes and titles, are tagged as speaker None. No annotation guidelines are provided, and it is not clarified in the paper how instances of quotation that are associated with a character, but are not spoken (such as written materials like letters) were handled in the annotation process.\footnote{It is fair to note that the haziness of Muzny et al.’s annotation categories is likely the result of the aim of the paper, which was not to identify and analyze dialogue per se, but rather to extract features most associated with ‘dialogism’ from dialogue, and to track those features’ use across the entire text of a set of novels. For this purpose, an extraction pipeline that results in mostly dialogue being extracted is sufficient.} The Quote Annotator test set is best described as a dataset of text, of all forms, enclosed in quotation marks in the corpus. However, the dataset can be used as a training/test set for direct speech and thought identification combined, by filtering out quotations tagged with no speaker.

\subsection{RiQuA}

The RiQuA dataset contains a combination of short stories annotated in full, and annotated samples from three full-length novels, tagged for direct and indirect speech, which the authors refer to as “speech acts”. \cite{papay-pado-2020-riqua}\footnote{The full dataset is available at https://www.ims.uni-stuttgart.de/en/research/resources/corpora/riqua} Two of the novels (Austen’s \textit{Emma} and Twain’s \textit{The Adventures of Tom Sawyer}), and five of the short stories (Dickens “A Christmas Carol”, and four of Arthur Conan Doyle’s Sherlock Holmes stories) are originally written in English. One of the novels (Flaubert’s \textit{Madame Bovary}, and three of the stories (all by Chekhov) are English translations. The texts and samples included are identical to those included in the Columbia Quoted Speech Attribution Corpus (Elson and McKeown, 2010), which annotated the same set of texts for speaker data, but extracted the quotations automatically. In addition to the manually tagged direct and indirect speech spans, this corpus also includes manual annotations for “cues” (speech tags like “said” and “replied”), speaker and addressee. The primary strengths of this dataset are its compatibility with the Columbia corpus, its mixture of genres (it is the only publicly available, hand-annotated direct discourse dataset that we could find that includes a substantive amount of short story data) and its inclusion of indirect speech data, which makes it the most suitable available dataset for researchers analyzing fictional speech more generally. Furthermore, the theoretical framing of RiQuA is the most concrete and consistent of the three available datasets, and Papay and Padó share their full annotation guidelines.

\section{NightmareDiscourseBench} 

\subsection{Benchmark Description}

In our review of the three benchmark datasets, we identified three primary gaps. First, the novels contained in all of these datasets are overwhelmingly canonical novels, published in the nineteenth century, and are also biased towards realist texts. The available test sets therefore offer a sample that is formally normative, and offer few examples of non-canonical texts that are more likely to pose challenges for rules-based direct discourse identification. While lower accuracy rates for novels that are deliberately experimental in their form may not be a source of concern for research that analyzes direct discourse trends at scale (because experimental novels are, by definition, outliers) the canonical texts that dominate the existing test sets risk obscuring the diverse forms in which direct discourse manifests across much larger, more varied corpora that include a substantial number of non-canonical novels. In essence, there is a space between the canonical and the explicitly experimental, in which direct discourse may be represented quite frequently in ways that challenge existing heuristic pipelines, which are not made visible in existing test sets. Pipelines tested on existing test sets are also likely to obscure potential issues in eighteenth or twentieth century texts, which often deviate from the formal norms of the nineteenth century. The Quote Annotator test set is transparent about the issue of period specificity—Muzny et al.’s pipeline performed dramatically worse on their benchmark’s only eighteenth-century text, Emmeline the Orphan.

When selecting texts for our benchmark, we therefore aimed to include several texts that:

\begin{enumerate}
    \item Are formally experimental, unusual or complex 
    \item Exhibit above-average use of techniques that increase error rates in existing direct discourse pipelines, such as embedded texts, scare quotes, dialect rendered using contractions, embedded direct discourse and long speeches that cross multiple paragraphs.
    \item Contain framing techniques like embedded narrators and epistolary novels
\end{enumerate}

In addition to selecting novels that contain challenges to rules-based direct discourse identification, we also manually selected samples from those novels that were particularly complex in this domain. Finally, we also selected texts that exhibited a range of typographical conventions for the demarcation of direct discourse–single straight quotes, single curly quotes, double straight quotes, and double curly quotes. A summary of the texts from which we sampled for our benchmark dataset is provided in Table 1 below. The benchmark contains samples of approximately 3000 words from 12 unique novels, with 12 unique authors. 

NightmareDiscourseBench also includes more extensive metadata than existing benchmarks. Our metadata aims to help researchers understand how certain variables affect automated pipelines. We include metadata for:

\begin{itemize}
  \setlength{\itemsep}{0pt}
  \item Title
  \item Author
  \item Year of publication
  \item Place of publication
  \item Quotation mark type
  \item Perspective (First person, third person or epistolary)
\end{itemize}

Finally, for each text we also provide Project Gutenberg source data and commentary on the features that make the sample challenging from a direct discourse perspective. A full list of texts and their associated metadata is provided in our GitHub repository for the dataset. \footnote{https://github.com/aih-lab/nightmare\_discourse\_bench} An abbreviated metadata overview for NightmareDiscourseBench is provided in Table 1 below. Table 2 summarizes the available metadata for LitBank, Quote Annotator, RiQuA and NightmareDiscourseBench.

\begin{table}[h]
\centering 
\resizebox{\textwidth}{!}{
\begin{tabular}{cccccc}
\toprule
Title & Author & Date & Place & Quote type & Narrator \\
\midrule
\textit{Little Women} & Louisa May Alcott & 1868 & US & double curly & third person \\
\textit{Alice in Wonderland} & Lewis Carroll & 1865 & UK & double curly & third person \\
\textit{Frankenstein} & Mary Shelley & 1818 & UK & double curly & epistolary \\
\textit{Ruth Hall} & Fanny Fern & 1854 & US & double straight & third person \\
\textit{Dream Life} & Donald Grant Mitchell & 1851 & US & double straight & first AND second person! \\
\textit{Clotel} & William Wells Brown & 1853 & US & double straight & third person \\
\textit{Our Mutual Friend} & Charles Dickens & 1865 & UK & single curly & third person \\
\textit{Emmeline, The Orphan of the Castle} & Charlotte Smith & 1788 & UK & single straight & third person \\
\textit{Pamela} & Samuel Richardson & 1740 & UK & none & epistolary \\
\textit{The Wild Irish Girl}\tablefootnote{\textit{The Wild Irish Girl}, by Anglo-Irish novelist Sydney Owenson, was first published in London but focuses its theme and settings on the question of Irish independence. While we list it as a UK novel here because our place data aims to help researchers understand how typographical norms in different publishing regions can affect the representation of direct discourse, the first author would like to acknowledge the place of the text in the Irish literary canon and the contested colonial status of Ireland at the time of its publication.} & Sydney Owenson (Lady Morgan) & 1806 & UK & double curly & epistolary \\
\textit{Unfettered: A Novel} & Sutton Griggs & 1902 & US & double straight & third person \\
\textit{Lorna Doone} & R. D. Blackmore & 1869 & UK & single straight & first person \\
\bottomrule
\end{tabular}
}
\caption{Summary metadata for NightmareDiscourseBench}
\label{tab:example}
\end{table}

\begin{table}[h]
\centering 
\begin{tabular}{cccccccc}
\toprule
 & Title & Author & Year & Source & Nation & Perspective & Typography \\
\midrule
LitBank & \checkmark & \checkmark & \checkmark & \checkmark & $\times$ & $\times$ & $\times$ \\
Quote Annotator & \checkmark & \checkmark & \checkmark & $\times$ & $\times$ & $\times$ & $\times$ \\
RiQuA & \checkmark & \checkmark & \checkmark & \checkmark & $\times$ & $\times$ & $\times$ \\
NDB & \checkmark & \checkmark & \checkmark & \checkmark & \checkmark & \checkmark & \checkmark \\
\bottomrule
\end{tabular}
\caption{Metadata available for existing test sets and NightmareDiscourseBench}
\label{tab:dataset-metadata}
\end{table}

\subsection{Annotation Process}

Our annotation process consisted of three rounds of annotation performed by two undergraduate research assistants. The annotators were familiar with the goals of the project, having worked with our research team on the project the preceding six months. Annotations were performed using a purpose-built annotation tool that allowed the annotators to highlight segments of text and tag them with one or more of our annotation categories. The tool outputs the results in the form of a structured JSON file. The authors and annotators first developed the annotation categories and guidelines iteratively by annotating an example text (chosen because it contained examples of several edge-cases for our categories, such as nested speech, scare quotes and extensive use of embedded text) and discussing any issues with the guidelines that arose from that process. After the annotation categories and guidelines were revised based on this initial practice and discussion, we developed a final set of annotation guidelines that included four categories: direct speech, direct thought, embedded text, and other. The other category includes edge-cases, the discourse status of which is ambiguous within the text, including scare quotes. Once this initial training process was complete, our annotators independently completed tagging the full 12 sample corpus. The annotation instructions used by our annotators, and the associated category descriptions and examples, are available in the Appendix (Figure 3).

The final set of annotations were created by first including annotations for which the same text span was tagged with the same category label. Then, one of the authors, a specialist in narrative theory, reviewed the two subsets of conflicting annotations–one in which a text span was tagged by both annotators, but with different category labels, and one of texts that were only tagged by one of the annotators. In these conflicting cases, the final determination for inclusion in the benchmark, and with what label, was made by this third expert annotator.  The annotations produced by our two annotators had a mean agreement rate of 96\% for tagged text spans. Seven of our texts had 100\% agreement for tagged text spans. The text with the lowest agreement rate was Pamela, with 79.5\% agreement for tagged text spans. We observe that this is due to Pamela’s epistolary format, and its lack of quotation marks to delimit direct speech. For tags, texts tagged by both annotators had a minimum one label mean agreement rate of 99\%. The texts tagged by both annotators had an exact label match (i.e., an identical set of labels for a given text were applied by all annotators) of 88\%. We observe that the majority of cases in which annotators shared one label, but disagreed on additional labels, was in cases of nested direct speech in which annotators were split between speech and other as a secondary label. We provide the full set of annotations, including tagged text spans, primary label, secondary labels, and source text in our GitHub repository for the project. In the interests of transparency, we also provide each individual annotator's initial annotations (before expert resolution) in full. A summary of the final benchmark dataset, along with overview statistics for the three existing test sets, is provided in Table 3 below. 

The most striking difference between our benchmark and the three benchmarks reviewed above is the longer mean length of our tagged text spans–because we deliberately chose text samples with features that often confound direct speech identification pipelines, our dataset contains more long text spans like passages of direct speech that run over several paragraphs, letters and embedded text like newspaper articles. Direct speech still dominates the benchmark however–556 text spans, close to 90\%, were tagged as speech. The benchmark contains 34 instances of embedded text, 15 instances of direct thought, and 16 text spans tagged as ‘other’.

\begin{table}[h]
\centering 
\resizebox{\textwidth}{!}{
\begin{tabular}{lccccc}
\toprule
 & LitBank & Quote Annotator & RiQuA & NightmareDiscourseBench \\
\midrule
Annotation Category & Dialogue & Quoted Text & Direct and indirect speech & Direct discourse (four subcategory labels) \\
No. of Texts & 86 & 6 & 10 & 12 \\
No. of Unique Authors & 75 & 6 & 6 & 12 \\
Partial or full novel annotation? & Partial (2000 tokens) & Full & Mixed & Partial ($\sim$3000 tokens) \\
No. of Tokens & 210,532 & 708,224 & 408,092 & 43,773 \\
No. of Tagged Tokens & 29186 & 257927 & 116707 & 23,860 \\
Total No. of Tagged Text Spans & 1765 & 13661 & 5963 & 621 \\
Rel. Frequency of Tagged Text Spans (per 1000 tokens) & 8.383 & 19.289 & 14.612 & 14.187 \\
Mean Length of Tagged Text Spans & 16.54 & 19.03 & 19.57 & 38.42 \\
SD in length of Tagged Spans & 28.57 & 40.28 & 44.62 & 154.04 \\
\bottomrule
\end{tabular}%
}
\caption{Descriptive Statistics for Existing Benchmarks and NightmareDiscourseBench}
\label{tab:dataset-summary}
\end{table}

\section{Comparative Testing with BookNLP}

In order to demonstrate the challenges posed by our benchmark dataset for direct discourse identification pipelines that rely on quotation-mark-based extraction, we run the most widely used tool for this task, BookNLP, on each of the available test datasets described in section 3 and NightmareDiscourseBench. BookNLP is a modular text processing pipeline designed for fiction that includes tools for tokenization, NER, part-of-speech tagging and several other NLP tasks. \cite{bamman2021booknlp} (We select BookNLP to evaluate our benchmark’s performance as a stress test for direct speech identification because it is a very popular tool for processing fictional texts. However, we wish to be clear that this is not intended to be a critique of BookNLP, which is an extensive, multi-purpose NLP pipeline.) BookNLP includes a speaker identification module which “identifies all instances of direct speech in the text and attributes it to its speaker.” \cite{bamman2021booknlp} The BookNLP direct speech and speaker identification module relies on identifying text enclosed by quotation marks to identify direct speech. Its authors report a 90.8 F1 score for BookNLP’s direct speech identification on a test subset of the LitBank dataset. \cite{sims2020measuring}

Evaluating accuracy for direct discourse identification is challenging. Accuracy calculated using exact token or character matches can produce misleading low scores, because of stray whitespaces or encoding issues that produce different punctuation or character formatting. Furthermore, different test datasets can have different formatting decisions (LitBank, for example, includes whitespace padding around text enclosed in quotation marks). Therefore, instead of evaluating BookNLP’s performance on each dataset using exact string matches, we use indel similarity scoring with a threshold of 0.9 to determine whether a string extracted by BookNLP has a match in each of our test datasets. A threshold of 0.9 maintains a high standard for string matching while allowing for minor formatting or data processing errors. Our code for running BookNLP on each test dataset and calculating accuracy is available in our GitHub for the project. The accuracy, reported as mean F1, precision and recall scores, and associated standard deviations, across all of the sample texts in each corpus, are provided in Table 4 below. We provide full accuracy data for each dataset in our accompanying GitHub repo. 

\begin{table}[h]
\centering 
\begin{tabular}{ccccc}
\toprule
 & LitBank & Quote Annotator & RiQuA & NDB \\
\midrule
Mean F1 & 0.88 & 0.74 & 0.85 & 0.77 \\
Mean Precision & 0.83 & 0.74 & 0.86 & 0.75 \\
Mean Recall & 0.93 & 0.74 & 0.83 & 0.79 \\
F1 SD & 0.38 & 0.24 & 0.16 & 0.28 \\
Precision SD & 0.38 & 0.23 & 0.16 & 0.28 \\
Recall SD & 0.41 & 0.25 & 0.18 & 0.27 \\
\bottomrule
\end{tabular}
\caption{BookNLP performance on direct discourse datasets}
\label{tab:example}
\end{table}

Surprisingly, BookNLP’s performance on NightmareDiscourseBench is, on average, slightly higher than the Quote Annotator benchmark, as demonstrated in Figure 1 below. However, the variation in F1 scores is more pronounced for NightmareDiscourseBench, as shown in Figure 2 below. Quote Annotator’s texts cluster into two groups–an upper bound cluster of three on which BookNLP performs very well, and a second cluster of three on which BookNLP performs in the 0.5-0.6 range. Nightmare DiscourseBench, in contrast, includes more texts in the intermediate 0.6-0.8 range. Overall, while NightmareDiscourseBench is more suitable as a benchmark than Quote Annotator because of its additional category labels and metadata, we recommend using it as a test set in tandem with Quote Annotator if stress testing direct speech NLP pipelines.

\begin{figure}[t!]
  \centering
  \includegraphics[width=0.8\linewidth]{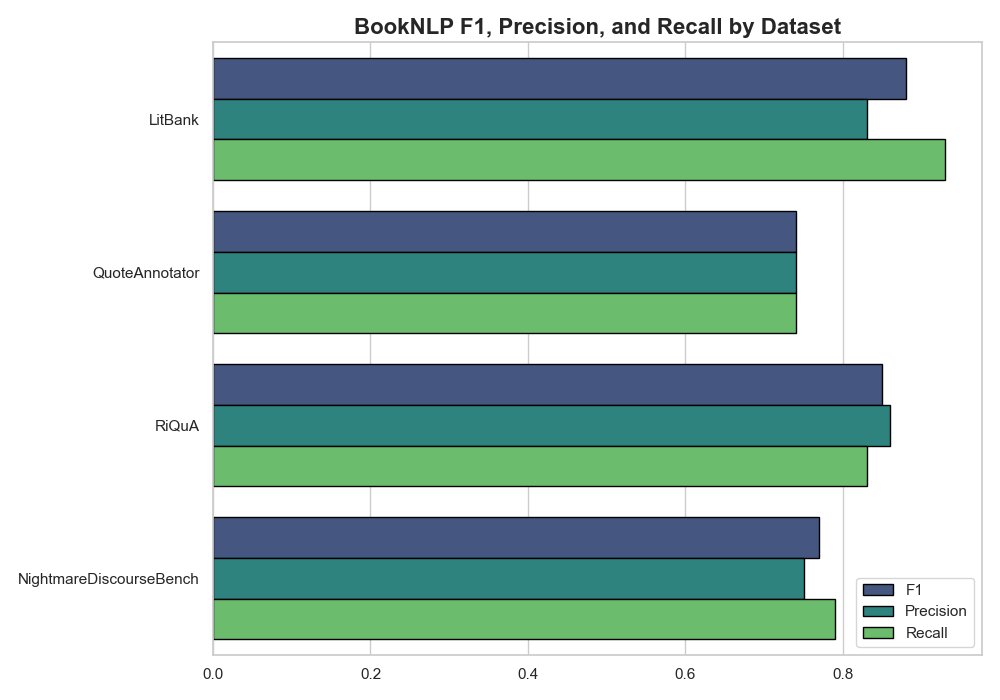}
  \caption{BookNLP F1, Precision and Recall scores on each dataset}
  \label{fig:example}
\end{figure}

\begin{figure}[t!]
  \centering
  \includegraphics[width=0.8\linewidth]{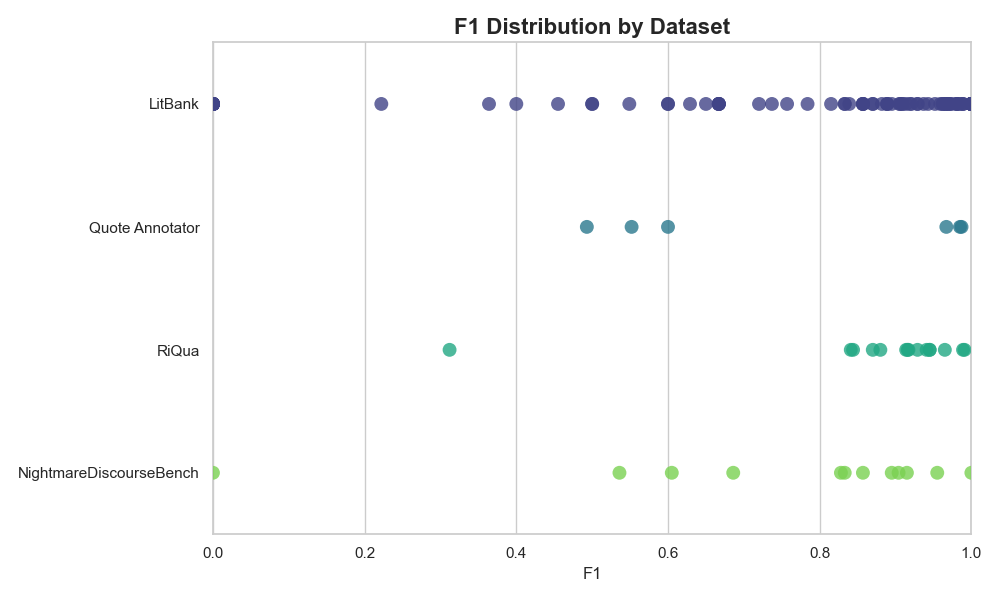}
  \caption{F1 Score distributions per dataset}
  \label{fig:example}
\end{figure}

\section{Limitations and Further Work} 

NightmareDiscourseBench’s primary weakness is its small size–because manual expert review is required at three levels (the text selection stage, the sample selection stage, and the annotation review stage) and double blind annotation is also necessary, assembling a large benchmark in this domain is a time-consuming process. An additional weakness of the benchmark, in common with the other three benchmarks reviewed here, is its concentration of nineteenth-century texts. This constraint is largely practical–because we wished to be able to provide full text sources for replication, we were restricted to texts in the public domain. We intend to iteratively continue building the benchmark in the coming years to make it more flexible for a wide range of researchers, with a focus on expanding the annotated data for direct thought and embedded text, and including more texts from the eighteenth and early twentieth centuries. 

The primary direction for further research is the development of an LLM-based NLP pipeline for this task that can handle differentiating different forms of direct discourse. While we test only on direct speech in this paper, because that is a category our own benchmark has in common with all three prior benchmarks, direct thought and embedded diegetic text like letters have their own representational affordances. A pipeline that can accurately identify such text would be of use to scholars working on topics such as the representation of interiority, communication networks and levels of narrative embedment in fiction. We are currently in the process of developing an LLM-based approach to this task, which we intend to have flexibility across different categories of direct discourse, as well as its different representational techniques.

\section{Acknowledgements}

This project was supported by the Digital Transformations Research Corp program at Washington University in St. Louis.

\printbibliography

@incollection{mchale2009,
  author = {McHale, Brian},
  title = {Speech Representation},
  booktitle = {Handbook of Narratology},
  editor = {Hühn, Peter and others},
  publisher = {De Gruyter},
  year = {2009},
  pages = {434--446},
  doi = {10.1515/9783110217445.434}
}

@book{prince,
    author = {Prince, Gerald},
    title = {A Dictionary of Narratology},
    publisher = {University of Nebraska Press},
    year = {1989}
}

@book{richardson2004,
    author = {Richardson, Samuel},
    title = {Pamela, or Virtue Rewarded},
    publisher = {Project Gutenberg},
    year = {2004},
    url = {https://www.gutenberg.org/cache/epub/6124/pg6124.txt}
}

@article{Muzny2017a,
  title  = "Dialogism in the novel: A computational model of the dialogic nature of narration and quotations",
  author    = "Muzny, Felix and Algee-Hewitt, Mark and Jurafsky, Dan",
  journal   = "Digital Scholarship in the Humanities",
  publisher = "Oxford University Press (OUP)",
  volume    =  32,
  number    = "suppl\_2",
  pages     = "ii31--ii52",
  month     =  dec,
  year      =  2017,
  language  = "en",
  doi = {10.1093/llc/fqx031}
}

@inproceedings{papay-pado-2020-riqua,
    title = "{RiQuA}: A Corpus of Rich Quotation Annotation for {English} Literary Text",
    author = "Papay, Sean  and
      Pad{\'o}, Sebastian",
    editor = "Calzolari, Nicoletta  and
      B{\'e}chet, Fr{\'e}d{\'e}ric  and
      Blache, Philippe  and
      Choukri, Khalid  and
      Cieri, Christopher  and
      Declerck, Thierry  and
      Goggi, Sara  and
      Isahara, Hitoshi  and
      Maegaard, Bente  and
      Mariani, Joseph  and
      Mazo, H{\'e}l{\`e}ne  and
      Moreno, Asuncion  and
      Odijk, Jan  and
      Piperidis, Stelios",
    booktitle = "Proceedings of the Twelfth Language Resources and Evaluation Conference",
    month = may,
    year = "2020",
    address = "Marseille, France",
    publisher = "European Language Resources Association",
    url = "https://aclanthology.org/2020.lrec-1.104/",
    pages = "835--841",
    language = "eng",
    isbn = "979-10-95546-34-4"}

@inproceedings{Jannidis2018AnalysingDS,
  title={Analysing Direct Speech in German Novels},
  author={Jannidis, Fotis and Konle, Leonard and Zehe, Albin and Hotho, Andreas and Krug, Markus},
  booktitle={Jahrestagung des Verbands Digital Humanities im deutschsprachigen Raum},
  year={2018},
  url={https://api.semanticscholar.org/CorpusID:235190470}
}

@book{wharton,
    author = {Wharton, Edith},
    title = {The Age of Innocence},
    publisher = {Project Gutenberg},
    year = {1996}
}

@article{menon2019,
  author = {Menon, T.},
  title = {Keeping Count: Direct Speech in the Nineteenth-Century British Novel},
  journal = {Narrative},
  year = {2019},
  volume = {27},
  number = {2},
  pages = {160--181},
  doi = {10.1353/nar.2019.0010}
}

@inproceedings{vishnubhotla2024emotion,
  author = {Vishnubhotla, Krishnapriya and Hammond, Adam and Hirst, Graeme and Mohammad, Saif},
  title = {The Emotion Dynamics of Literary Novels},
  booktitle = {Findings of the Association for Computational Linguistics: ACL 2024},
  year = {2024},
  pages = {2557--2574},
  address = {Bangkok, Thailand},
  publisher = {Association for Computational Linguistics},
  doi = {10.18653/v1/2024.findings-acl.150}
}

@article{lee2024conversational,
  author = {Lee, John S. Y. and Webster, Carol},
  title = {Conversational networks: prophets and kings in the Old Testament},
  journal = {Digital Scholarship in the Humanities},
  year = {2024},
  volume = {39},
  number = {4},
  pages = {1043--1063},
  doi = {10.1093/llc/fqae049}
}

@inproceedings{sims2020measuring,
  author = {Sims, Matthew and Bamman, David},
  title = {Measuring Information Propagation in Literary Social Networks},
  booktitle = {Proceedings of the 2020 Conference on Empirical Methods in Natural Language Processing (EMNLP)},
  year = {2020},
  pages = {642--652},
  address = {Online},
  publisher = {Association for Computational Linguistics},
  doi = {10.18653/v1/2020.emnlp-main.47}
}

@article{kubis2021quantitative,
  author = {Kubis, Marek},
  title = {Quantitative analysis of character networks in Polish 19th- and 20th-century novels},
  journal = {Digital Scholarship in the Humanities},
  year = {2021},
  volume = {36},
  number = {suppl\_2},
  pages = {ii175--ii181},
  doi = {10.1093/llc/fqab012}
}

@inproceedings{elson2010extracting,
  author = {Elson, David and Dames, Nicholas and McKeown, Kathleen},
  title = {Extracting Social Networks from Literary Fiction},
  booktitle = {Proceedings of the 48th Annual Meeting of the Association for Computational Linguistics},
  year = {2010},
  pages = {138--147},
  address = {Uppsala, Sweden},
  publisher = {Association for Computational Linguistics}
}

@inproceedings{he2013identification,
  author = {He, Hua and Barbosa, Denilson and Kondrak, Grzegorz},
  title = {Identification of Speakers in Novels},
  booktitle = {Proceedings of the 51st Annual Meeting of the Association for Computational Linguistics (Volume 1: Long Papers)},
  year = {2013},
  pages = {1312--1320},
  address = {Sofia, Bulgaria},
  publisher = {Association for Computational Linguistics}
}

@inproceedings{michel2024improving,
  author = {Michel, Gaspard and Epure, Elena V. and Hennequin, Romain and Cerisara, Christophe},
  title = {Improving Quotation Attribution with Fictional Character Embeddings},
  booktitle = {Findings of the Association for Computational Linguistics: EMNLP 2024},
  year = {2024},
  pages = {12723--12735},
  address = {Miami, Florida, USA},
  publisher = {Association for Computational Linguistics},
  doi = {10.18653/v1/2024.findings-emnlp.744}
}

@inproceedings{michel2024distinguishing,
  author = {Michel, Gaspard and Epure, Elena and Hennequin, Romain and Cerisara, Christophe},
  title = {Distinguishing Fictional Voices: a Study of Authorship Verification Models for Quotation Attribution},
  booktitle = {Proceedings of the 8th Joint SIGHUM Workshop on Computational Linguistics for Cultural Heritage, Social Sciences, Humanities and Literature (LaTeCH-CLfL 2024)},
  year = {2024},
  pages = {160--171},
  address = {St. Julians, Malta},
  publisher = {Association for Computational Linguistics},
  doi = {10.18653/v1/2024.latechclfl-1.15}
}

@inproceedings{byszuk2020detecting,
  author = {Byszuk, Joanna and Woźniak, Michał and Kestemont, Mike and Leśniak, Albert and Łukasik, Wojciech and {\v{S}e\c{l}a}, Artjoms and Eder, Maciej},
  title = {Detecting Direct Speech in Multilingual Collection of 19th-century Novels},
  booktitle = {Proceedings of LT4HALA 2020 -- 1st Workshop on Language Technologies for Historical and Ancient Languages},
  year = {2020},
  pages = {100--104},
  address = {Marseille, France},
  publisher = {European Language Resources Association (ELRA)}
}

@inproceedings{kurfali2020zeroshot,
  author = {Kurfalı, Murathan and Wirén, Mats},
  title = {Zero-shot cross-lingual identification of direct speech using distant supervision},
  booktitle = {Proceedings of the 4th Joint SIGHUM Workshop on Computational Linguistics for Cultural Heritage, Social Sciences, Humanities and Literature},
  year = {2020},
  pages = {105--111},
  address = {Online},
  publisher = {International Committee on Computational Linguistics}
}

@inproceedings{allaith2025annotating,
  author = {Al-Laith, Ali and Conroy, Alexander and Degn, Kirstine Nielsen and Bjerring-Hansen, Jens and Hershcovich, Daniel},
  title = {Annotating and Classifying Direct Speech in Historical Danish and Norwegian Literary Texts},
  booktitle = {Proceedings of the Joint 25th Nordic Conference on Computational Linguistics and 11th Baltic Conference on Human Language Technologies (NoDaLiDa/Baltic-HLT 2025)},
  year = {2025},
  pages = {1--7},
  address = {Tallinn, Estonia},
  publisher = {University of Tartu Library}
}

@inproceedings{stymne2024direct,
  author = {Stymne, Sara},
  title = {Direct Speech Identification in Swedish Literature and an Exploration of Training Data Type, Typographical Markers, and Evaluation Granularity},
  booktitle = {Proceedings of the 8th Joint SIGHUM Workshop on Computational Linguistics for Cultural Heritage, Social Sciences, Humanities and Literature (LaTeCH-CLfL 2024)},
  year = {2024},
  pages = {253--263},
  address = {St. Julians, Malta},
  publisher = {Association for Computational Linguistics},
  doi = {10.18653/v1/2024.latechclfl-1.25}
}

@inproceedings{stymne2022slanda,
  author = {Stymne, Sara and Östman, Carin},
  title = {{SLäNDa} version 2.0: Improved and Extended Annotation of Narrative and Dialogue in Swedish Literature},
  booktitle = {Proceedings of the Thirteenth Language Resources and Evaluation Conference},
  year = {2022},
  pages = {5324--5333},
  address = {Marseille, France},
  publisher = {European Language Resources Association}
}

@article{brunner2013automatic,
  author = {Brunner, Annelen},
  title = {Automatic recognition of speech, thought, and writing representation in German narrative texts},
  journal = {Literary and Linguistic Computing},
  year = {2013},
  volume = {28},
  number = {4},
  pages = {563--575},
  doi = {10.1093/llc/fqt024}
}

@inproceedings{schoch2016straight,
  author = {Schöch, C. and Schlör, D. and Popp, S. and Brunner, A. and Henny, U. and Calvo Tello, J.},
  title = {Straight Talk! Automatic Recognition of Direct Speech in Nineteenth-Century French Novels},
  booktitle = {Digital Humanities 2016: Conference Abstracts},
  year = {2016},
  pages = {346--353},
  address = {Kraków},
  publisher = {Jagiellonian University and Pedagogical University}
}

@misc{bamman2019litbank,
  author = {Bamman, David},
  title = {{LitBank}},
  year = {2019},
  howpublished = {GitHub},
  url = {https://github.com/dbamman/litbank}
}

@misc{bamman2021booknlp,
  author = {Bamman, David},
  title = {{BookNLP}},
  year = {2021},
  howpublished = {GitHub},
  url = {https://github.com/booknlp/booknlp}
}

@inproceedings{lee2016annotated,
  author = {Lee, John and Yeung, Chak Yan},
  title = {An Annotated Corpus of Direct Speech},
  booktitle = {Proceedings of the Tenth International Conference on Language Resources and Evaluation (LREC'16)},
  year = {2016},
  pages = {1059--1063},
  address = {Portorož, Slovenia},
  publisher = {European Language Resources Association (ELRA)}
}

@article{hess2020narrator,
  author = {Hess, Leopold and Bary, Corien},
  title = {Narrator language and character language in Thucydides: A quantitative study of narrative perspective},
  journal = {Digital Scholarship in the Humanities},
  year = {2020},
  volume = {35},
  number = {3},
  pages = {557--586},
  doi = {10.1093/llc/fqz026}
}

@inproceedings{vishnubhotla2022dialogism,
  author = {Vishnubhotla, Krishnapriya and Hammond, Adam and Hirst, Graeme},
  title = {The {Project Dialogism Novel Corpus}: A Dataset for Quotation Attribution in Literary Texts},
  booktitle = {Proceedings of the Thirteenth Language Resources and Evaluation Conference},
  year = {2022},
  pages = {5838--5848},
  address = {Marseille, France},
  publisher = {European Language Resources Association}
}

@article{brooke2017lexical,
  author = {Brooke, Julian and Hammond, Adam and Hirst, Graeme},
  title = {Using models of lexical style to quantify free indirect discourse in modernist fiction},
  journal = {Digital Scholarship in the Humanities},
  year = {2017},
  volume = {32},
  number = {2},
  pages = {234--250},
  doi = {10.1093/llc/fqv072}
}
\appendix

\clearpage
\section{Additional Materials}
\label{appdx:first}

\begin{longtable}{@{}p{0.23\textwidth}p{0.72\textwidth}@{}}

\toprule
\textbf{Challenge} & \textbf{Example} \\
\midrule
\endfirsthead

\toprule
\textbf{Challenge} & \textbf{Example} \\
\midrule
\endhead

\endfoot

\bottomrule
\noalign{\vskip 0.3cm}
\caption{Examples of Challenges in Dialogue Identification}
\label{tab:dialogue-challenges}\\
\endlastfoot

Scare quotes &
No one ever regretted the admittance of Sam Weller, for a more devoted, well-behaved, and jovial member no club could have. He certainly did add ‘spirit’ to the meetings, and ‘a tone’ to the paper, for his orations convulsed his hearers and his contributions were excellent, being patriotic, classical, comical, or dramatic, but never sentimental. \\

\midrule

Titles/Names of works, organizations or businesses &
This mine host of the Atlantic promised me, pointing to a little board suspended over the door, on which was written:

\par\medskip
“Good Dry Lodging.”
\par\medskip

My landlord, however, convinced me his hotel afforded something better than good dry lodging; for entreating me to alight, till a shower passed over which was beginning to fall, I entered the hut, and found his wife, a sturdy lad their eldest son, and two naked little ones, seated at their dinner, and enjoying such a feast, as Apicius, who sailed to Africa from Rome to eat good oysters, would gladly have voyaged from Rome to Ireland to have partaken of; for they were absolutely dining on an immense turbot (whose fellow-sufferers were floundering in a boat that lay anchored near the door.) \\

\midrule

Quotations &
Wearied by my restlessness, rather than refreshed by my transient slumbers, I arose with the dawn, and carrying my \emph{port-feuille} and pencils with me, descended from my tower, and continued to wander for some time among the wild and romantic scenes which surround these interesting ruins, while

\par\medskip
“La sainte recueihnent la paisible innocence

\par\medskip
Sembler de ces lieus habiter le silence.”
\par\medskip

until almost wearied in the contemplation of the varying sublimities which the changes of the morning’s seasons shed over the ocean’s boundless expanse, from the first gray vapour that arose from its swelling wave, to that splendid refulgence with which the risen sun crimsoned its bosom, I turned away my dazzled eye, and fixed it on the ruins of Inismore. \\

\midrule

Contractions &
'Latt the young ozebird aloun, zay I. Makk zuch ado about un, wi' hogs'-puddens, and hock-bits, and lambs'-mate, and whaten bradd indade, and brewers' ale avore dinner-time, and her not to zit wi' no winder aupen--draive me mad 'e doo, the ov'ee, zuch a passel of voouls. Do 'un good to starve a bit; and takk zome on's wackedness out ov un.' \\

\midrule

Multi-paragraph speech &
‘Some twelve days I lived at that hotel, before the night when I saw the corpse of Radfoot at the Police Station. The inexpressible mental horror that I laboured under, as one of the consequences of the poison, makes the interval seem greatly longer, but I know it cannot have been longer. That suffering has gradually weakened and weakened since, and has only come upon me by starts, and I hope I am free from it now; but even now, I have sometimes to think, constrain myself, and stop before speaking, or I could not say the words I want to say.

\par\medskip
‘Again I ramble away from thinking it out to the end. It is not so far to the end that I need be tempted to break off. Now, on straight! \\

\midrule

Absent or non-standard typographical markers &
And so, as I was saying, he had these clothes come home, and he tried them on. And before he pulled them off, he sent for me, when nobody else was in the parlour with him: Pamela, said he, you are so neat and so nice in your own dress, (Alack-a-day, I didn’t know I was!) that you must be a judge of ours. How are these clothes made? Do they fit me?—I am no judge, said I, and please your honour; but I think they look very fine. \\

\end{longtable}

\par\medskip

\noindent
\begin{minipage}{\linewidth}
    \centering

    \includegraphics[
        width=\linewidth,
        height=0.22\textheight,
        keepaspectratio
    ]{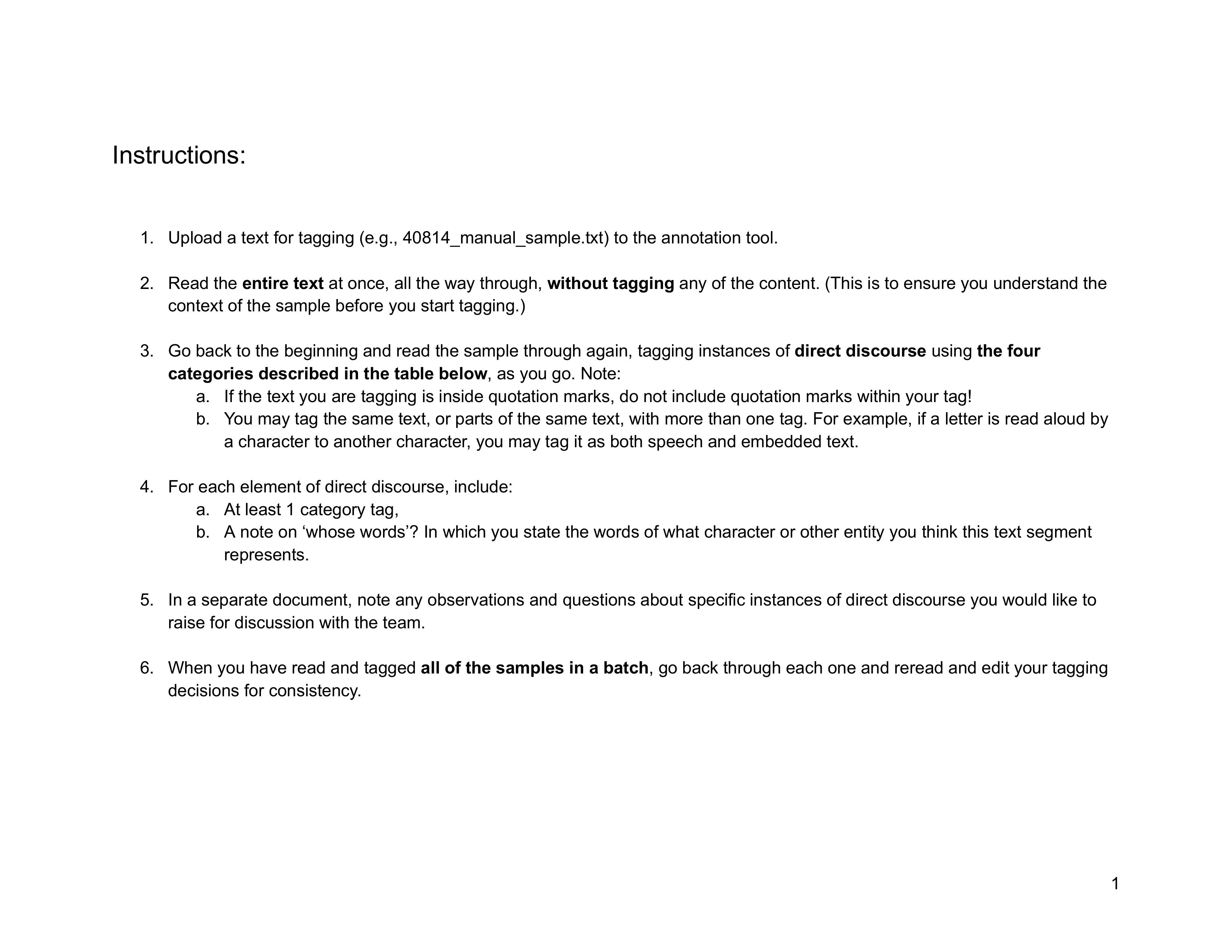}

    \par\vspace{0.2cm}

    \includegraphics[
        width=\linewidth,
        height=0.22\textheight,
        keepaspectratio
    ]{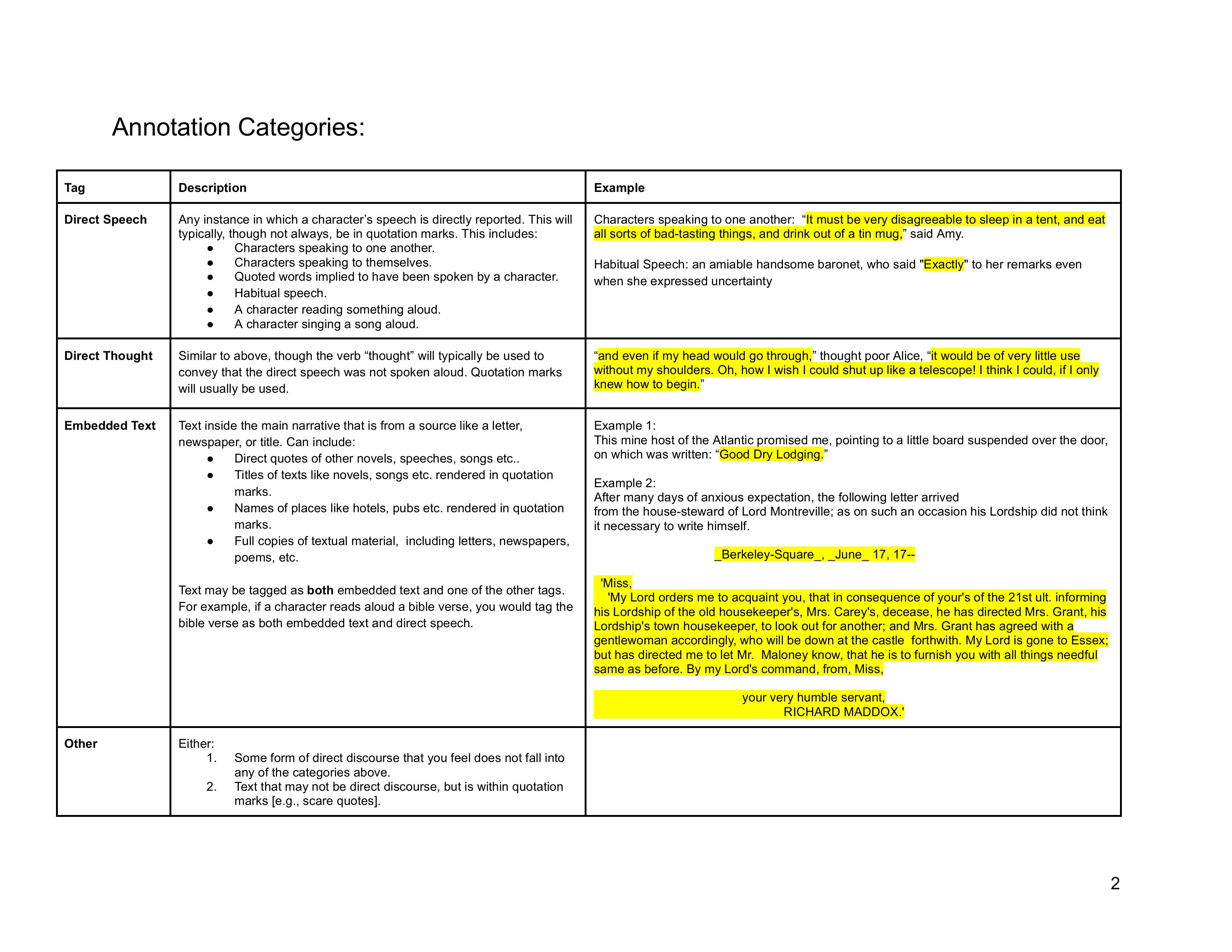}

    \par\vspace{0.2cm}

    \includegraphics[
        width=\linewidth,
        height=0.22\textheight,
        keepaspectratio
    ]{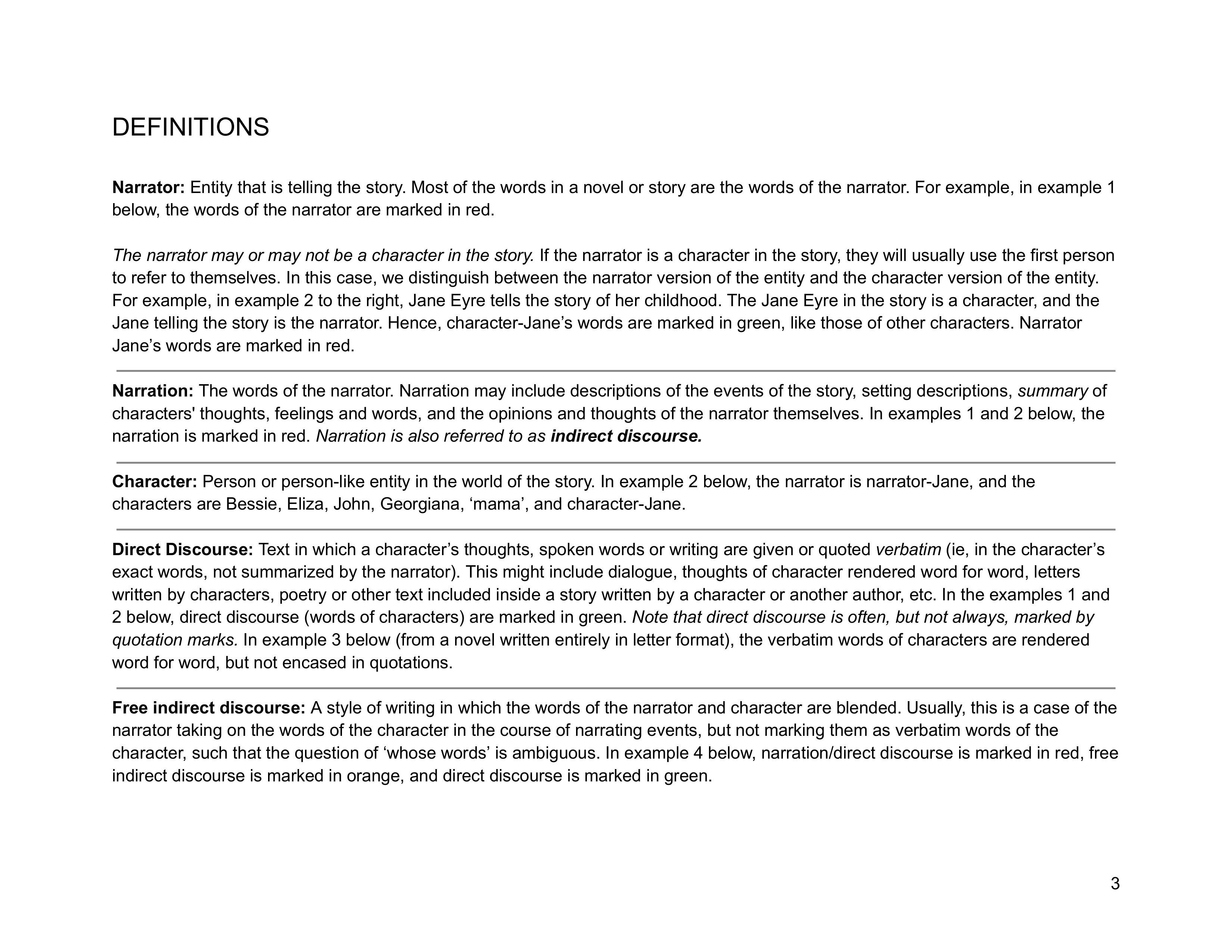}

    \captionof{figure}{Annotation guidelines used by student annotators}
    \label{fig:annotation-guidelines}

\end{minipage}

\end{document}